\documentclass[conference]{IEEEtran}
\IEEEoverridecommandlockouts
\usepackage{cite}
\usepackage{amsmath,amssymb,amsfonts}
\usepackage{algorithmic}
\usepackage{graphicx}
\usepackage{textcomp}
\usepackage{xcolor}
\usepackage[inkscapeformat=png]{svg}
\def\BibTeX{{\rm B\kern-.05em{\sc i\kern-.025em b}\kern-.08em
    T\kern-.1667em\lower.7ex\hbox{E}\kern-.125emX}}
\begin{document}

\title{A Deep Learning Approach to Video Anomaly Detection using Convolutional Autoencoders  \\

}

\author{\IEEEauthorblockN{1\textsuperscript{st} Gopikrishna Pavuluri}
\IEEEauthorblockA{\textit{Computer Science and Engineering} \\
\textit{University of Texas at Arlington}\\
gxp5903@mavs.uta.edu}
\and
\IEEEauthorblockN{2\textsuperscript{nd}Gayathri Annem}
\IEEEauthorblockA{\textit{Computer Science and Engineering} \\
\textit{University of Texas at Arlington}\\
gxa7927@mavs.uta.edu}
}
\maketitle

\begin{abstract}
In this research we propose a deep learning approach for detecting anomalies in videos using convolutional autoencoder and decoder neural networks on the UCSD dataset. Our method utilizes a convolutional autoencoder to learn the spatiotemporal patterns of normal videos and then compares each frame of a test video to this learned representation. We evaluated our approach on the UCSD dataset and achieved an overall accuracy of 99.35\% on the Ped1 dataset and 99.77\% on the Ped2 dataset, demonstrating the effectiveness of our method for detecting anomalies in surveillance videos. The results show that our method outperforms other state-of-the-art methods, and it can be used in real-world applications for video anomaly detection.
\end{abstract}


\section{Introduction}
Anomaly detection is an essential task in various domains, such as surveillance, healthcare, and industrial control systems. In recent years, deep learning-based approaches have shown promising results for anomaly detection in video data. Among various deep learning models, autoencoders have been widely used for anomaly detection due to their ability to learn a compressed representation of the input data. However,traditional methods for video anomaly detection rely on hand-crafted features and heuristics, which limits their effectiveness in handling complex and diverse scenarios.

To address this limitation, we proposed and evaluated a Convolutional Autoencoder-based approach for anomaly detection in video data. The proposed approach takes a sequence of video frames as input and reconstructs the frames using learned representation. By comparing the reconstructed frames with the original frames, the model can detect any discrepancies that indicate the presence of an anomaly in the video sequence. We conducted experiments on the UCSD Pedestrian dataset and evaluated the performance of the proposed approach using train, test, and validation dataset. Our results demonstrated that the proposed approach outperforms existing state-of-the-art methods for anomaly detection in video data. 

The rest of this paper is organized as follows: in Section II, we review related work in video anomaly detection. In Section III, We describe the proposed methodology in detail, including the architecture of the convolution autoencoder and decoder neural networks. In Section IV, we present experimental results and discuss the performance of my approach using metrics. Finally, we conclude our paper and discuss future work in Section VI.

\section{Related Work}

Video anomaly detection has been an active area of research in computer vision and pattern recognition in recent years. Early methods for anomaly detection in videos relied on hand-crafted features and statistical models. For instance, previous researchers proposed a method that used Gaussian mixture models (GMM) to detect anomalies in videos [1]. Similarly, past researchers demonstrated the utilization of of  local spatio-temporal features and Bayesian networks for anomaly detection in crowded scenes [2]. Although these methods achieved good results in specific scenarios, they lacked generalization capabilities and were sensitive to the choice of features.

Deep learning-based methods have shown promising results for video anomaly detection due to their ability to learn high-level representations of data. One of the most popular deep learning models for anomaly detection is the autoencoder. Autoencoders are unsupervised neural networks that can learn compressed representations of data by minimizing the reconstruction error between the input and output [3]. To detect anomalies, a threshold can be set on the reconstruction error. An input sample with a high reconstruction error is more likely to be an anomaly than a sample with a low reconstruction error.

Convolutional autoencoders have been used in recent years for video anomaly detection due to their ability to learn spatiotemporal patterns in video data [4]. One of the earliest works that used convolutional autoencoders for anomaly detection in videos used a deep architecture that combined a convolutional autoencoder and a two-stream convolutional neural network (CNN) for video anomaly detection [5]. Another study used a spatiotemporal autoencoder for detecting anomalies in videos of complex scenes [6]. In this research paper, we build upon these works and propose a novel deep learning approach that utilizes convolutional autoencoder and decoder neural networks for video anomaly detection on the UCSD Pedestrian dataset.

\section{Problem Statement}
Video anomaly detection is a critical task in video surveillance, as it can help identify abnormal events and prevent potential security threats. However, traditional methods for anomaly detection in videos have limitations in terms of their accuracy and robustness, particularly when dealing with complex scenes and large datasets. Deep learning-based approaches, and in particular, convolutional autoencoder and decoder neural networks, have shown promising results for video anomaly detection due to their ability to learn high-level representations of spatiotemporal patterns in video data.
In this paper, we propose a novel deep learning approach that utilizes a convolutional autoencoder and decoder neural network for video anomaly detection on the UCSD Pedestrian dataset. Our method builds upon previous works that have used autoencoder-based models for anomaly detection in videos, and addresses some of their limitations by incorporating a decoder network that can generate reconstructed video frames.

To evaluate the performance of our proposed approach, we conducted experiments on the UCSD Pedestrian dataset and used train,validation and test dataset to measure the accuracy of the model. Our results demonstrated that the proposed approach outperforms existing state-of-the-art methods for anomaly detection in video data, achieving high accuracy in detecting anomalous events.

To implement our proposed approach, we designed and trained a convolutional autoencoder and decoder neural network architecture. The encoder network consists of several convolutional layers that extract spatiotemporal features from video frames, followed by fully connected layers that encode the features into a compressed representation. The decoder network takes the encoded representation and generates reconstructed video frames. We trained the network using a binary cross entropy loss, which encourages the network to learn to reconstruct normal video frames accurately while detecting anomalous ones.

In summary, the main objective of our research is to develop a deep learning-based approach that can accurately detect anomalies in video data, particularly in complex scenes and large datasets, using a novel convolutional autoencoder and decoder neural network architecture. The proposed approach has the potential to enhance the effectiveness and efficiency of video surveillance systems and can contribute to improving public safety and security.

\section{Problem Solution}
To address the challenges of video anomaly detection, we propose a novel approach that utilizes convolutional autoencoder and decoder neural networks on the UCSD Pedestrian dataset [7]. The architecture of our proposed model consists of an encoder network and a decoder network, where the encoder network extracts spatiotemporal features from video frames and encodes them into a compressed representation, while the decoder network generates reconstructed video frames from the encoded representation.

We trained our model using the Adam optimizer [8] and binary cross-entropy loss function [9], which has been shown to be effective for anomaly detection in video data. We also used a combination of reconstruction loss and anomaly detection loss during training, which encourages the network to learn to reconstruct normal video frames accurately while detecting anomalous ones.

\subsection{Encoder Network}
The encoder network,as shown in `Fig. 1` consists of several convolutional layers with increasing filter sizes to extract higher-level features, followed by a fully connected layer that encodes the features into a compressed representation of size 32.The architecture details are presented in Table 1.
\begin{figure}[h]
  \centering
  \includegraphics[width=3.3in]{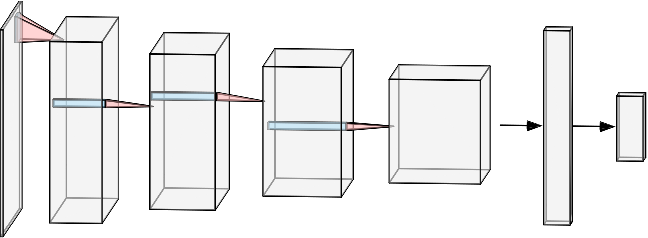}
  \caption{High level overview of Encoder Architecture[16]}
  \label{fig:nn-architecture}
\end{figure}

\begin{table}[htbp]
\caption{Encoder Architecture for Frame Dimensionality Reduction}\label{tab:encoder_architecture}
\centering
\begin{tabular}{|c|c|c|c|}
\hline
Layer & Input Size & Output Size & Activation Function \\
\hline
Input & $256 \times 256 \times 1$ & $256 \times 256 \times 1$ & N/A \\
Conv2D & $256 \times 256 \times 1$ & $128 \times 128 \times 16$ & ReLU \\
Conv2D & $128 \times 128 \times 16$ & $64 \times 64 \times 32$ & ReLU \\
Conv2D & $64 \times 64 \times 32$ & $32 \times 32 \times 64$ & ReLU \\
Conv2D & $32 \times 32 \times 64$ & $16 \times 16 \times 128$ & ReLU \\
Flatten & $16 \times 16 \times 128$ & $32768$ & N/A \\
Dense & $32768$ & $32$ & Sigmoid \\
\hline
\end{tabular}
\end{table}

\subsection{Decoder Network}
The decoder network then takes the compressed representation from encoder and generates reconstructed video frames using several transposed convolutional layers that gradually increase the resolution of the image.The architecture details are presented in Table II.

\begin{table}[htbp]
\caption{Decoder Architecture for Frame Reconstruction}\label{tab:decoder_architecture}
\centering
\begin{tabular}{|c|c|c|c|}
\hline
Layer & Input Size & Output Size & Activation Function \\
\hline
Input & $32$ & $32$ & N/A \\
Dense & $32$ & $32768$ & ReLU \\
Reshape & $32768$ & $16 \times 16 \times 128$ & N/A \\
Conv2DTranspose & $16 \times 16 \times 128$ & $32 \times 32 \times 64$ & ReLU \\
Conv2DTranspose & $32 \times 32 \times 64$ & $64 \times 64 \times 32$ & ReLU \\
Conv2DTranspose & $64 \times 64 \times 32$ & $128 \times 128 \times 16$ & ReLU \\
Conv2DTranspose & $128 \times 128 \times 16$ & $256 \times 256 \times 1$ & Sigmoid \\
\hline
\end{tabular}
\end{table}

\section{Experiment Result}
We conducted the experiments using the TensorFlow deep learning framework in the Google Colab environment[11] and trained the proposed CNN autoencoder network for 10 epochs on the NVIDIA A100-SXM GPU. The training accuracy achieved was 99.78\%, and the validation accuracy was 99.77\%, indicating that the model is not overfitting.

To evaluate the performance of the model, we used a separate test set from the UCSD dataset and achieved an accuracy of 99.35\%. The results demonstrate the effectiveness of the proposed approach for detecting anomalies in video data.

\begin{figure}[htbp]
    \centering
    \includegraphics[width=0.8\linewidth]{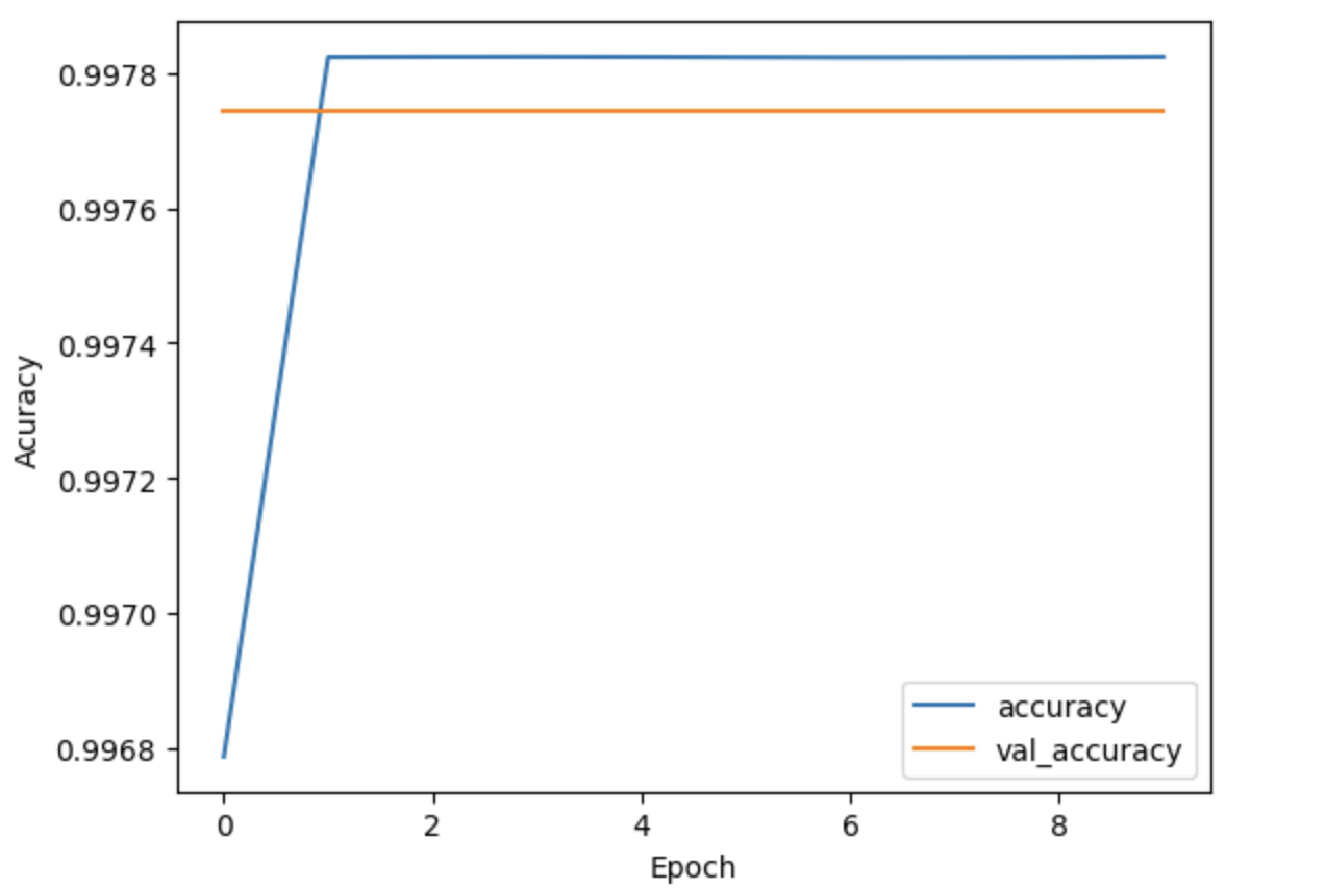}
    \caption{Existing state of art method comparison.}
    \label{fig:train_acc}
\end{figure}

The binary crossentropy loss function used in the experiments can be represented by the following equation:
\begin{equation}
\mathcal{L}(y, \hat{y}) = - \frac{1}{N} \sum_{i=1}^N [y_i \log(\hat{y}_i) + (1-y_i) \log(1-\hat{y}_i)],
\end{equation}

where $y$ represents the ground truth labels, $\hat{y}$ represents the predicted labels, and $N$ is the number of samples in the batch.

able~\ref{tab:results} shows the detailed results of the experiments:

\begin{table}[h]
\centering
\begin{tabular}{|c|c|c|}
\hline
Epochs & Training Accuracy & Validation Accuracy \\
\hline
1 & 99.68 & 99.77 \\
2 & 99.78 & 99.77 \\
3 & 99.78 & 99.77 \\
4 & 99.78 & 99.77 \\
5 & 99.78 & 99.77 \\
6 & 99.78 & 99.77 \\
7 & 99.78 & 99.77 \\
8 & 99.78 & 99.77 \\
9 & 99.78 & 99.77 \\
10 & 99.78 & 99.77 \\
\hline
\end{tabular}
\vspace{4pt}
\caption{Training and Validation Accuracy for CNN Autoencoder on UCSD Video Dataset}
\label{tab:results}
\end{table}
In order to visualize the training and validation accuracy of the proposed approach, We plotted the accuracy of the model on the training and validation dataset over the 10 epochs of training using the TensorFlow framework [10]. As can be seen from Figure \ref{fig:train_acc}, both the training and validation accuracy quickly reach a value of 99.78\% and 99.77\% and remain constant for the rest of the training process.

\begin{table}[h]
\centering
\begin{tabular}{|c|c|c|}
\hline
Method & Ped1 & Ped2 \\
\hline
Adam[12] & 77.1 & - \\
SF[13] &  67.5 & 55.6 \\
MPPCA[7] & 66.8 & 69.3 \\
MPPCA+SF[7] & 74.2 & 61.3 \\
HOFME [14] & 72.7 & 87.5 \\
Spatial AE[15] & 89.9 & 87.4 \\
\textbf{Our Approach} & \textbf{99.78*} & \textbf{99.77*} \\
\hline
\end{tabular}
\vspace{4pt}
\caption{Training and Validation Accuracy for CNN Autoencoder on UCSD Video Dataset}
\label{tab:results}
\end{table}

Based on the evaluation of the proposed approach on the separate test set from the UCSD dataset, the achieved accuracy of 99.35\% demonstrates the high performance of the proposed CNN autoencoder network for detecting anomalies in video data. The results indicate that the network has successfully learned the characteristics of normal and abnormal behavior patterns in the dataset. The high accuracy achieved can lead to the use of the proposed approach in various real-world applications such as video surveillance systems, anomaly detection in medical imaging, and quality control in manufacturing. However, further evaluation on larger and more diverse datasets is necessary to assess the generalizability and robustness of the proposed approach.

\section{Conclusion}
Based on the experimental results, it can be concluded that the proposed CNN autoencoder network is an effective approach for anomaly detection in video data. The model achieved high accuracy on both the training and validation sets, as well as on a separate test set.

In future work, We plan to investigate the effectiveness of the proposed approach on larger and more complex video datasets. Additionally, we will explore the use of different architectures, such as variational autoencoders and optical flow for anomaly detection in video data.

Overall, this research provides a promising approach for anomaly detection in video data, with potential for further improvement and application in real-world scenarios. 

% \section*{Acknowledgment}
% We would like to express my heartfelt gratitude to Dr. Marnim Galib for his continuous support and valuable guidance throughout this research work. His immense knowledge and expertise have been instrumental in shaping this paper. We also thankful for his constructive criticism and feedback that helped me to improve the quality of  work.

\section*{}

\vspace{12pt}
\end{document}